\documentclass[runningheads]{llncs}

\usepackage{eccv}

\usepackage{eccvabbrv}

\usepackage{graphicx}
\usepackage{booktabs}

\usepackage[accsupp]{axessibility}  

\usepackage[pagebackref,breaklinks,colorlinks,citecolor=eccvblue]{hyperref}

\usepackage{orcidlink}

\usepackage{multirow} 
\usepackage{pifont}

\begin{document}

\title{Video Object Segmentation via SAM 2: The 4th Solution for LSVOS Challenge VOS Track} 

\titlerunning{The 4th Solution for LSVOS Challenge VOS Track}

\author{Feiyu Pan \and
Hao Fang \and
Runmin Cong \and
Wei Zhang \and
Xiankai Lu}

\authorrunning{F.~Pan et al.}

\institute{Shandong University\\
Team: MVP-TIME}

\maketitle

\begin{abstract}
Video Object Segmentation (VOS) task aims to segmenting a particular object instance throughout the entire video sequence given only the object mask of the first frame. Recently, Segment Anything Model 2 (SAM 2) is proposed, which is a foundation model towards solving promptable visual segmentation in images and videos. SAM 2 builds a data engine, which improves model and data via user interaction, to collect the largest video segmentation dataset to date. SAM 2 is a simple transformer architecture with streaming memory for real-time video processing, which trained on the date provides strong performance across a wide range of tasks. In this work, we evaluate the zero-shot performance of SAM 2 on the more challenging VOS datasets MOSE and LVOS. Without fine-tuning on the training set, SAM 2 achieved 75.79 \( \mathcal{J} \)\&\( \mathcal{F} \) on the test set and ranked 4th place for 6th LSVOS Challenge VOS Track.
  \keywords{Video object segmentation \and Segment anything}
\end{abstract}

\section{Introduction}
\label{sec:intro}
Unlike Unsupervised Video Object Segmentation (VOS)~\cite{lu2020zero,lu2021segmenting}, Semi-supervised VOS usually begins with an object mask as input in the first frame, which must be accurately tracked throughout the video. This task has drawn significant attention due to its relevance in various applications, including robotics~\cite{petrik2022learning}, video editing~\cite{cheng2021modular}, reducing costs in data annotation~\cite{athar2023burst}. 
The 6th LSVOS challenge features two tracks: the VOS track and the RVOS track. This year, the challenge has introduced new datasets, MOSE~\cite{ding2023mose} and MeViS~\cite{ding2023mevis}, replacing the classic YouTube-VOS~\cite{xu2018youtube} and Referring YouTube-VOS benchmarks~\cite{seo2020urvos} used in previous editions. The new datasets present more complex scenes, including scenarios with disappearing and reappearing objects, inconspicuous small objects, heavy occlusions, crowded environments, and long-term videos, making this year's challenge more difficult than ever before.

Early neural network-based approaches have often used online fine-tuning on the first video frame~\cite{caelles2017one,robinson2020learning} or on all frames ~\cite{voigtlaender2017online} to adapt the model to the target object. However, fine-tuning is slow during test-time. Recent VOS approaches employ a memory-based paradigm~\cite{oh2019video,cheng2022xmem,yang2021associating}. A memory representation is computed from past segmented frames, and any new query frame “reads” from this memory to retrieve features for segmentation. But pixel-level matching lacks high-level consistency and is prone to matching noise, especially in the presence of distractors. To address this, Cutie~\cite{cheng2024putting} proposes object-level memory reading, which effectively puts the object from a memory back into the query frame. Inspired by recent query-based video segmentation~\cite{fang2024learning,fang2024unified} that represent objects as “object queries”, Cutie implements object-level memory reading with an object transformer, achieving the state-of-the-art performance among specialized VOS models.

Segment Anything (SA)~\cite{kirillov2023segment} introduced a foundation model for promptable segmentation in images. While SA successfully addresses segmentation in images, existing video segmentation models and datasets fall short in providing a comparable capability to “segment anything in videos”. Recently, Meta present Segment Anything Model 2 (SAM 2)~\cite{ravi2024sam}, a foundation model towards solving promptable visual segmentation in images and videos. SAM 2 builds a data engine, which improves model and data via user interaction, to collect the largest video segmentation dataset to date. SAM 2 is a simple transformer architecture with streaming memory for real-time video processing. SAM 2 trained on the data provides strong performance across a wide range of tasks.

In this work, we evaluate the zero-shot performance of SAM 2 on the more challenging VOS datasets MOSE and LVOS. Without fine-tuning on the training set, SAM 2 achieved 75.79 \( \mathcal{J} \)\&\( \mathcal{F} \) on the test set and ranked 4th place for LSVOS Challenge VOS Track.

\section{Method}
\label{sec:metho}
\begin{figure}[t]
\begin{center}
\includegraphics[width=1\linewidth]{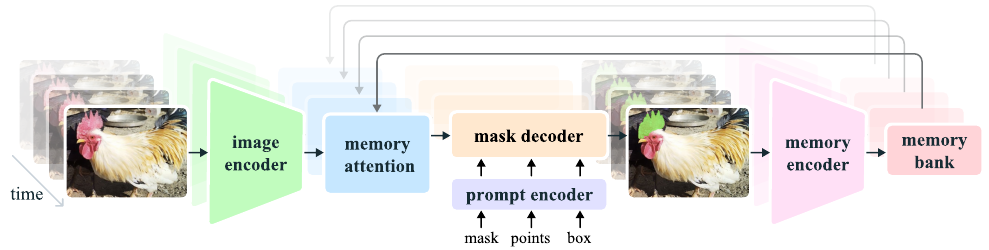}
\end{center}
\caption{The SAM 2 architecture. For a given frame, the segmentation prediction is conditioned on the current prompt and/or on previously observed memories. Videos are processed in a streaming fashion with frames being consumed one at a time by the image encoder, and cross-attended to memories of the target object from previous frames. The mask decoder, which optionally also takes input prompts, predicts the segmentation mask for that frame. Finally, a memory encoder transforms the prediction and image encoder embeddings (not shown in the figure) for use in future frames.}
\label{fig:model}
\end{figure}

As shown in ~\cref{fig:model}, SAM 2 supports point, box, and mask prompts on individual frames to define the spatial extent of the object to be segmented across the video. For image input, the model behaves similarly to SAM. A promptable and light-weight mask decoder accepts a frame embedding and prompts on the current frame and outputs a segmentation mask for the frame. Prompts can be iteratively added on a frame in order to refine the masks. Unlike SAM, the frame embedding used by the SAM 2 decoder is not directly from an image encoder and is instead conditioned on memories of past predictions and prompted frames. It is possible for prompted frames to also come “from the future” relative to the current frame. Memories of frames are created by the memory encoder based on the current prediction and placed in a memory bank for use in subsequent frames. The memory attention operation takes the per-frame embedding from the image encoder and conditions it on the memory bank to produce an embedding that is then passed to the mask decoder.

\section{Experiment}
\label{sec:exper}
\subsection{Datasets and Metrics}
\noindent \textbf{Datasets.} 
The dataset for this challenge is a mixture of MOSE~\cite{ding2023mose} and LVOS~\cite{hong2023lvos} datasets. MOSE~\cite{ding2023mose} contains 2,149 video clips and 5,200 objects from 36 categories, with 431,725 high-quality object segmentation masks. The most notable feature of MOSE dataset is complex scenes with crowded and occluded objects. The target objects in the videos are commonly occluded by others and disappear in some frames. LVOS~\cite{hong2023lvos} is the first densely annotated long-term VOS dataset, which consists of 220 videos with a total duration of 421 minutes. Each video includes various attributes, especially challenges deriving from the wild, such as long-term reappearing and cross-temporal similar objects.

\noindent \textbf{Evaluation Metrics.} 
we employ region similarity $\mathcal{J}$ (average IoU), contour accuracy $\mathcal{F}$ (mean boundary similarity), and their average \( \mathcal{J} \)\&\( \mathcal{F} \) as our evaluation metrics.

\subsection{Implementation Details}
We first test Cuite~\cite{cheng2024putting} using the official Cutie-base+ weight provided. This version is trained on the MEGA, which is the aggregated dataset consisting of DAVIS~\cite{perazzi2016benchmark}, YouTubeVOS~\cite{xu2018youtube}, MOSE~\cite{ding2023mose}, OVIS~\cite{qi2022occluded}, and BURST~\cite{athar2023burst}. To make Cutie more adaptable to the challenge dataset, we finetune the challenge training set on the weight. We use the AdamW optimizer with a learning rate of 1e-4, a batch size of 16, and a weight decay of 0.001. The training lasts for 20K iterations, with the learning rate reduced by 10 times after 15K iterations. We finally test SAM 2~\cite{ravi2024sam} using the officially provided Hiera-L weight. The weight is trained on SA-V manual + Internal and SA-1B.

\begin{figure}[t]
\begin{center}
\includegraphics[width=1\linewidth]{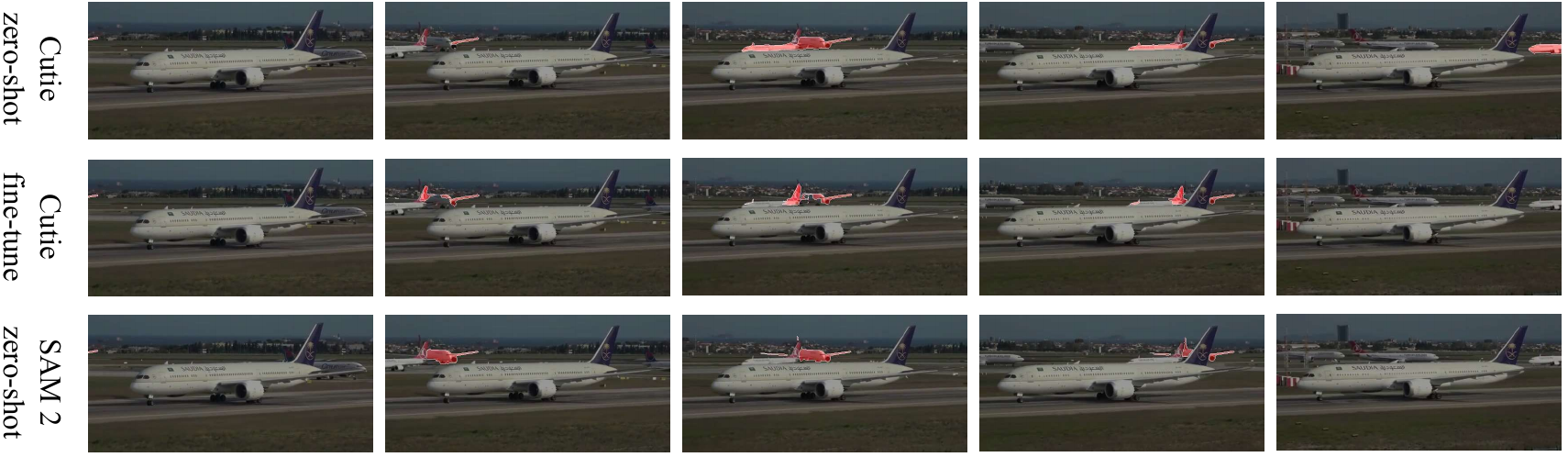}
\end{center}
\caption{Qualitative comparison of Cutie and SAM 2 on validation dataset.}
\label{fig:vos}
\end{figure}

\subsection{Main Results}
As shown in ~\cref{tab:val}, SAM 2 achieves 73.89\% in \( \mathcal{J} \)\&\( \mathcal{F} \), not only 2 points higher than Cutie, but even 1 point higher than the fine-tuned Cutie. ~\cref{fig:vos} shows some visualization results, where SAM 2 accurately segments and tracks the specified aircraft. As shown in ~\cref{tab:test}, SAM 2 achieves 75.79\% in \( \mathcal{J} \)\&\( \mathcal{F} \), ranking in 4th place in the 6th LSVOS Challenge RVOS Track at ECCV 2024.
\begin{table}
  \setlength\tabcolsep{10pt}
  \centering
  \caption{The performance comparison on the VOS Track val set.}
  \begin{tabular}{l|ccc}
    \toprule
    Method & \( \mathcal{J} \)\&\( \mathcal{F} \) & $\mathcal{J}$ & $\mathcal{F}$ \\
    \midrule
    Cutie & 71.90 & 68.56 & 75.24\\
    Cutie-finetune & 72.88 & 69.22 & 76.54\\
    \bf SAM 2 & \bf 73.89 & \bf 70.19 & \bf 77.59\\
  \bottomrule
\end{tabular}
\label{tab:val}
\end{table}

\begin{table}
  \setlength\tabcolsep{10pt}
  \centering
  \caption{The leaderboard of the VOS Track test set.}
  \begin{tabular}{l|ccc}
    \toprule
    Team & \( \mathcal{J} \)\&\( \mathcal{F} \) & $\mathcal{J}$ & $\mathcal{F}$ \\
    \midrule
    PCL VisionLab & 80.90 & 76.16 & 85.63\\
    yuanjie & 80.84 & 76.42 & 85.26\\
    Xy-unu & 79.52 & 75.16 & 83.88\\
    \bf MVP-TIME (Ours) & 75.79 & 71.25 & 80.33\\
  \bottomrule
\end{tabular}
\label{tab:test}
\end{table}

\section{Conclusion}
In this work, we evaluate the zero-shot performance of SAM 2 on the more challenging VOS datasets MOSE and LVOS. We also finetune and test the performance of Cutie to compare with SAM 2. Without fine-tuning on the training set, SAM 2 achieved 75.79 \( \mathcal{J} \)\&\( \mathcal{F} \) on the test set and ranked 4th place for LSVOS Challenge VOS Track. The experimental results demonstrate the powerful zero-shot performance of SAM 2, providing a reference for future VOS applications using SAM 2.

\bibliographystyle{splncs04}
\bibliography{main}

\begin{thebibliography}{10}
\providecommand{\url}[1]{\texttt{#1}}
\providecommand{\urlprefix}{URL }
\providecommand{\doi}[1]{https://doi.org/#1}

\bibitem{athar2023burst}
Athar, A., Luiten, J., Voigtlaender, P., Khurana, T., Dave, A., Leibe, B., Ramanan, D.: Burst: A benchmark for unifying object recognition, segmentation and tracking in video. In: WACV. pp. 1674--1683 (2023)

\bibitem{caelles2017one}
Caelles, S., Maninis, K.K., Pont-Tuset, J., Leal-Taix{\'e}, L., Cremers, D., Van~Gool, L.: One-shot video object segmentation. In: CVPR. pp. 221--230 (2017)

\bibitem{cheng2024putting}
Cheng, H.K., Oh, S.W., Price, B., Lee, J.Y., Schwing, A.: Putting the object back into video object segmentation. In: CVPR. pp. 3151--3161 (2024)

\bibitem{cheng2022xmem}
Cheng, H.K., Schwing, A.G.: Xmem: Long-term video object segmentation with an atkinson-shiffrin memory model. In: ECCV. pp. 640--658. Springer (2022)

\bibitem{cheng2021modular}
Cheng, H.K., Tai, Y.W., Tang, C.K.: Modular interactive video object segmentation: Interaction-to-mask, propagation and difference-aware fusion. In: CVPR. pp. 5559--5568 (2021)

\bibitem{ding2023mevis}
Ding, H., Liu, C., He, S., Jiang, X., Loy, C.C.: Mevis: A large-scale benchmark for video segmentation with motion expressions. In: ICCV. pp. 2694--2703 (2023)

\bibitem{ding2023mose}
Ding, H., Liu, C., He, S., Jiang, X., Torr, P.H., Bai, S.: Mose: A new dataset for video object segmentation in complex scenes. In: ICCV. pp. 20224--20234 (2023)

\bibitem{fang2024unified}
Fang, H., Wu, P., Li, Y., Zhang, X., Lu, X.: Unified embedding alignment for open-vocabulary video instance segmentation. arXiv preprint arXiv:2407.07427  (2024)

\bibitem{fang2024learning}
Fang, H., Zhang, T., Zhou, X., Zhang, X.: Learning better video query with sam for video instance segmentation. TCSVT  (2024)

\bibitem{hong2023lvos}
Hong, L., Chen, W., Liu, Z., Zhang, W., Guo, P., Chen, Z., Zhang, W.: Lvos: A benchmark for long-term video object segmentation. In: ICCV. pp. 13480--13492 (2023)

\bibitem{kirillov2023segment}
Kirillov, A., Mintun, E., Ravi, N., Mao, H., Rolland, C., Gustafson, L., Xiao, T., Whitehead, S., Berg, A.C., Lo, W.Y., et~al.: Segment anything. In: ICCV. pp. 4015--4026 (2023)

\bibitem{lu2020zero}
Lu, X., Wang, W., Shen, J., Crandall, D., Luo, J.: Zero-shot video object segmentation with co-attention siamese networks. PAMI  \textbf{44}(4),  2228--2242 (2020)

\bibitem{lu2021segmenting}
Lu, X., Wang, W., Shen, J., Crandall, D.J., Van~Gool, L.: Segmenting objects from relational visual data. PAMI  \textbf{44}(11),  7885--7897 (2021)

\bibitem{oh2019video}
Oh, S.W., Lee, J.Y., Xu, N., Kim, S.J.: Video object segmentation using space-time memory networks. In: ICCV. pp. 9226--9235 (2019)

\bibitem{perazzi2016benchmark}
Perazzi, F., Pont-Tuset, J., McWilliams, B., Van~Gool, L., Gross, M., Sorkine-Hornung, A.: A benchmark dataset and evaluation methodology for video object segmentation. In: CVPR. pp. 724--732 (2016)

\bibitem{petrik2022learning}
Petr{\'\i}k, V., Qureshi, M.N., Sivic, J., Tapaswi, M.: Learning object manipulation skills from video via approximate differentiable physics. In: IROS. pp. 7375--7382. IEEE (2022)

\bibitem{qi2022occluded}
Qi, J., Gao, Y., Hu, Y., Wang, X., Liu, X., Bai, X., Belongie, S., Yuille, A., Torr, P.H., Bai, S.: Occluded video instance segmentation: A benchmark. IJCV  \textbf{130}(8),  2022--2039 (2022)

\bibitem{ravi2024sam}
Ravi, N., Gabeur, V., Hu, Y.T., Hu, R., Ryali, C., Ma, T., Khedr, H., R{\"a}dle, R., Rolland, C., Gustafson, L., et~al.: Sam 2: Segment anything in images and videos. arXiv preprint arXiv:2408.00714  (2024)

\bibitem{robinson2020learning}
Robinson, A., Lawin, F.J., Danelljan, M., Khan, F.S., Felsberg, M.: Learning fast and robust target models for video object segmentation. In: CVPR. pp. 7406--7415 (2020)

\bibitem{seo2020urvos}
Seo, S., Lee, J.Y., Han, B.: Urvos: Unified referring video object segmentation network with a large-scale benchmark. In: ECCV. pp. 208--223. Springer (2020)

\bibitem{voigtlaender2017online}
Voigtlaender, P., Leibe, B.: Online adaptation of convolutional neural networks for video object segmentation. arXiv preprint arXiv:1706.09364  (2017)

\bibitem{xu2018youtube}
Xu, N., Yang, L., Fan, Y., Yue, D., Liang, Y., Yang, J., Huang, T.: Youtube-vos: A large-scale video object segmentation benchmark. arXiv preprint arXiv:1809.03327  (2018)

\bibitem{yang2021associating}
Yang, Z., Wei, Y., Yang, Y.: Associating objects with transformers for video object segmentation. NeurIPS  \textbf{34},  2491--2502 (2021)

\end{thebibliography}

\end{document}